\documentclass[letterpaper, 10 pt, conference]{ieeeconf}  
\pdfoutput=1

\IEEEoverridecommandlockouts                              

\usepackage{graphics} 
\usepackage{subfigure}
\usepackage{epsfig} 
\usepackage{amsmath} 
\usepackage[T1]{fontenc}
\usepackage{bm}
\usepackage{cite}
\usepackage{amsfonts}
\usepackage{booktabs}
\usepackage{multirow}
\usepackage{diagbox}
\usepackage{threeparttable}
\usepackage{colortbl}
\usepackage{hyperref}
\usepackage{mathrsfs}
\hypersetup{
    colorlinks=false, 
    }
\hypersetup{hidelinks} 
\newcommand \transpose {\mathsf{T}} 

\title{\LARGE \bf
Learning to Estimate 3-D States of Deformable Linear Objects from Single-Frame Occluded Point Clouds
}

\author{Kangchen Lv, Mingrui Yu, Yifan Pu, Xin Jiang, Gao Huang, and Xiang Li
\thanks{K. Lv, M. Yu, Y. Pu, G. Huang, and X. Li are with the Department of Automation, Tsinghua University, China. X. Jiang is with the Beijing Academy of Artificial Intelligence, Beijing, China.
This work was supported in part by the National Key R\&D Program of China under Grant 2020AAA0105200, in part by the Institute for Guo Qiang, Tsinghua University, and in part by the National Natural Science Foundation of China under Grant U21A20517 and 52075290. Corresponding author: Xiang Li (xiangli@tsinghua.edu.cn)}}

\begin{document}

\maketitle
\thispagestyle{empty}
\pagestyle{empty}

\begin{abstract}
Accurately and robustly estimating the state of deformable linear objects (DLOs), such as ropes and wires, is crucial for DLO manipulation and other applications. However, it remains a challenging open issue due to the high dimensionality of the state space, frequent occlusions, and noises. 
This paper focuses on learning to robustly estimate the states of DLOs from single-frame point clouds in the presence of occlusions using a data-driven method. 
We propose a novel two-branch network architecture to exploit global and local information of input point cloud respectively and design a fusion module to effectively leverage the advantages of both methods. 
Simulation and real-world experimental results demonstrate that our method can generate globally smooth and locally precise DLO state estimation results even with heavily occluded point clouds, which can be directly applied to real-world robotic manipulation of DLOs in 3-D space.
\end{abstract}

\section{Introduction}
Robotic manipulation of deformable linear objects (DLOs), such as ropes and wires, has a wide variety of applications in industrial, service, and health-care sectors\cite{zhu2021challenges, yin2021modeling}. An accurate and robust state estimator for DLOs is obviously the prerequisite for subsequent manipulations. Compared to rigid objects, the infinite dimensional DLO state space makes it very challenging to perceive deformations. Besides, occlusions and noises occur frequently in unstructured environments, resulting in higher requirements for robust DLO state estimation.

\begin{figure} [tb]
  \centering 
    \includegraphics[width=8.5cm]{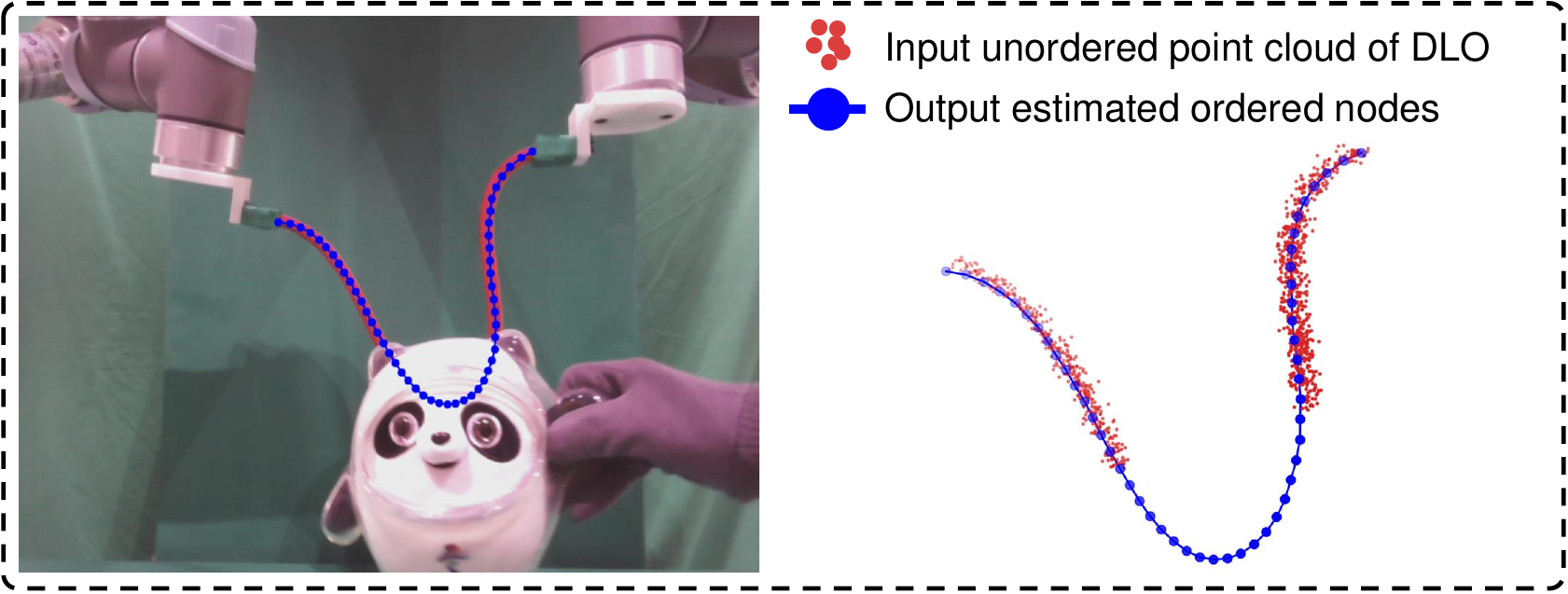} 
  \vspace{-2mm}
  \caption{Illustration of our task: 3-D occlusion-robust DLO state estimation from a single-frame point cloud. Red points are the unordered incomplete point cloud of the occluded rope and blue connected dots represent our estimated ordered node sequence as its current state.}
  \label{fig:first_figure}
    \vspace{-3mm}
\end{figure}

Commonly used representations to describe DLO states include Fourier-based parameterization\cite{navarro2017fourier}, implicit latent descriptors learned by neural networks\cite{sundaresan2020learning,zhou2021lasesom}, a chain of uniformly distributed nodes\cite{yu2022global,jin2019robust,yu2022shape,yu2022coarse}, etc. Among these methods, representing a DLO as a chain of 3-D nodes (see Fig. \ref{fig:first_figure}) is general in various  manipulation tasks and will be adopted in this work.


A complete processing stream to estimate the DLO state can be roughly divided into three procedures: segmentation (i.e., segmenting the DLO from environment), detection (i.e., estimating the DLO state in a single frame), and tracking (i.e., tracking the deformation across several frames).
As sensors are RGB or RGB-D cameras in most cases, segmenting the DLO region in image space is the essential first step for consecutive processing. 
\cite{dinkelwire,caporali2022fastdlo,gregorio2018let, zanella2021auto} focus on how to obtain pixel-level DLO masks of high quality using traditional image processing or data-driven methods. 
As for detection, this step aims at estimating the positions of nodes along the DLO in one frame with the  cleaned sensory data as input.
For example, \cite{huo2022keypoint,yan2020self} use neural networks to encode the DLO into several sequential key-points in the 2-D image space; \cite{wnuk2020kinematic} estimates a skeleton line and 3-D joint positions on it from point cloud to represent the DLO, but not robust against occlusions and different DLO types. 
As for tracking, various works have also been proposed to track the correspondence of point cloud across video frames in the presence of occlusions and self-intersections\cite{schulman2013tracking, tang2017state, tang2018track, tang2018framework, chi2019occlusion, wang2021tracking, jin2022robotic}. These works model DLO tracking task as a GMM-based non-rigid point registration problem with some geometric constraints. However, these pure tracking-based methods rely on an accurate initial state which requires manual setting or specific initial conditions. Besides, there are few effective ways to rectify the accumulated drift errors or re-initialize for tracking failure.
Therefore, it is necessary to develop an accurate and robust 3-D state estimation method for DLOs from a single frame, which can be independently applied to estimate the DLO state in each frame or combined with tracking methods above to utilize temporal information.

\begin{figure*} [tb]
  \centering 
    \includegraphics[width=0.85\textwidth]{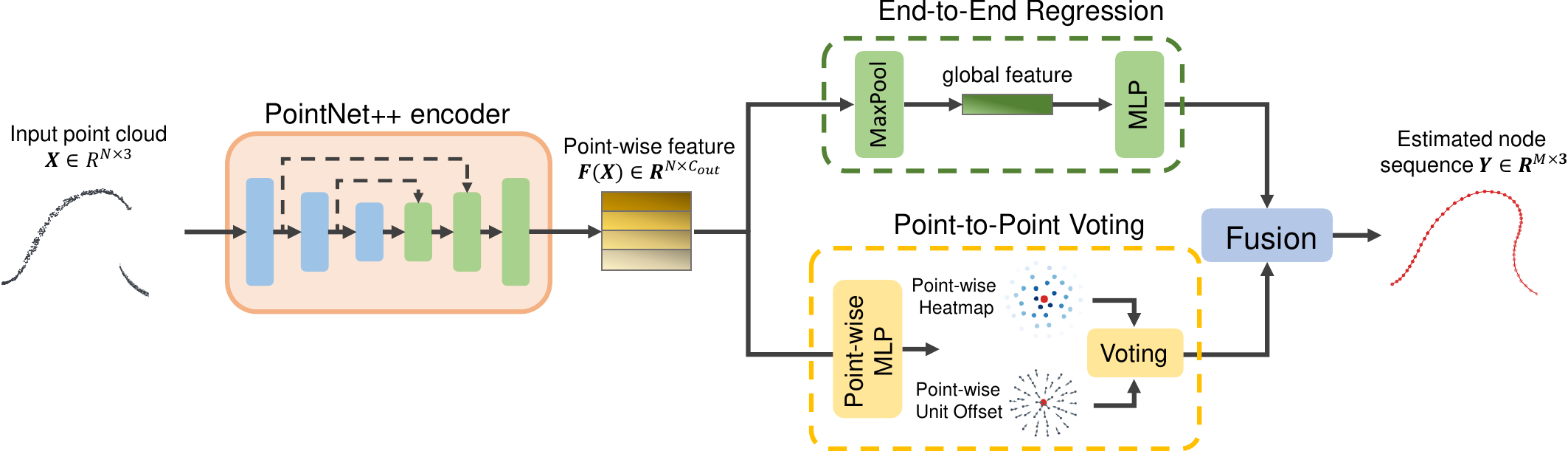} 
  \caption{Overview of the proposed method for occlusion-robustly estimating the 3-D states of DLOs. The input point cloud which might be fragmented due to occlusions is first fed into a PointNet++ encoder and the extracted features are then processed by two parallel branches: \textit{End-to-End Regression} and \textit{Point-to-Point Voting}. The estimation results of these two branches are finally fused with a fusion module to obtain the final output node sequence.}
  \label{fig:overview}
    \vspace{-2mm}
\end{figure*}

In this paper, we focus on estimating a sequence of ordered and uniformly distributed nodes from single-frame point cloud occlusion-robustly to represent the state of DLO, as shown in Fig. \ref{fig:first_figure}. Note that we only use point cloud as our input without any auxiliary physical simulation and robot configurations. The challenges of this task are as follows: 1) there are few distinguishable features in the point cloud of DLOs; 2) occlusions and noises are common in the environment; 3) generalization ability for different DLOs is required. To deal with challenges above, we propose a novel two-branch network architecture to leverage both the global geometry information for guaranteeing smooth and occlusion-robust shape, and local geometry information for precise estimations. To the best of our knowledge, we are the first to realize accurate and robust 3-D state estimation of DLOs from single-frame point cloud input even with heavy occlusions.
Specifically, we first exploit a PointNet++ encoder\cite{qi2017pointnet++} to extract deep features of the input point cloud and then feed the features into two branches: \textit{End-to-End Regression} and \textit{Point-to-Point Voting}. We encourage these two branches to focus on global and local geometry information respectively and finally fuse their estimations to combine their advantages. The whole framework is trained on synthetic dataset generated in simulation without collecting real-world data.
Experiments suggest our method achieves high performance on occlusion-robust state estimation of DLOs and can be directly applied in real-world scenarios.




\section{Problem Statement}
The goal of our method is to estimate the 3-D states of DLOs from point cloud obtained by an RGB-D camera. In this work, we focus on the state estimation problem and assume that the point cloud of the DLO has already been segmented out of the raw full point cloud by RGB image segmentation.
We represent the DLO state as a sequence of $M$ \textit{nodes} uniformly distributed, where $M$ is a pre-defined number of nodes that can sufficiently describe the DLO state.
The problem is to estimate the coordinates of the nodes $\bm Y=\left[\bm y_1, \bm y_2, \cdots , \bm y_M \right]^\transpose{} \in \mathbb{R}^{M\times 3}$ from the input point cloud $\bm X=\left[\bm x_1, \bm x_2, \cdots, \bm x_N \right]^\transpose{} \in \mathbb{R}^{N \times 3}$ where $N$ is the number of the points in the segmented point cloud.
Note that the input point cloud $\bm X$ is unordered, while the order of estimated nodes in $\bm Y$ from one end to another end has been represented by the index $1,2,\cdots,M$.
In addition, the point cloud of the DLO may be fragmentary and noisy because of the occlusions, imperfect segmentation, and depth images of low quality.

\section{Method}

As shown in Fig.~\ref{fig:overview}, our proposed method contains two branches: an \textit{End-to-End Regression} branch and a \textit{Point-to-Point Voting} branch, which focuses on the global and the local geometry information, respectively. 
Then, a deformable registration module is designed to leverage the advantages of both branches and fuse the two predictions to output the final estimated node sequence.

\subsection{End-to-End Regression}

The most straightforward approach is to train an end-to-end network with the point cloud $X\in \mathbb{R}^{N\times 3}$ as input and the node sequence $Y\in \mathbb{R}^{M\times 3}$ as output, which is indicated as \textit{End-to-End Regression}. 
We exploit a PointNet++\cite{qi2017pointnet++} encoder denoted as $\bm F(\cdot)$ to extract deep latent features $\bm F(\bm X)\in \mathbb{R}^{N\times C_{out}}$ of input point cloud $\bm X$, which means that each point in input point cloud has a $C_{out}$-dimensional feature vector.
A max pooling layer is then applied to get the global feature ${\rm MaxPool} ( \bm F(\bm X))\in \mathbb{R}^{C_{out}}$ which is irrelevant to the input point order. 
Finally, a fully-connected layer $\bm{FC}_1$ predicts the node sequence $\bm Y^{\rm pred}_{\rm reg}$. 
The whole regression network is defined as
\begin{equation}
    \bm Y^{\rm pred}_{\rm reg}= \bm{FC}_1({\rm MaxPool}(\bm{F}(\bm X))).
\end{equation}
With the ground-truth node coordinates $\bm Y^{\rm gt}$, the training loss function for each sample is
\begin{equation}
    L_{\rm reg}= \| \bm Y^{\rm pred}_{\rm reg} - \bm Y^{\rm gt} \|^2.
\end{equation}

It is experimentally found that such an end-to-end network can ensure that the estimated DLO shapes are smooth and look like real DLOs even using heavily occluded point cloud input, which suggests that this network can  learn the key global characteristic of DLOs well.
However, the predictions are often slightly different from the actual states such that they are not sufficiently accurate for applications (see Fig. \ref{fig:exp_sim_rvf}).
This phenomenon is believed to be brought about by the feature max pooling operation, which neglects crucial local information for precise estimation.

\subsection{Point-to-Point Voting}

To make up for the shortcomings of the end-to-end regression method, we design a point-to-point voting framework to utilize local geometry information, which is inspired by early works \cite{ge2018point, wan2018dense}.
Instead of using max-pooling layers for direct regression, this method generates point-wise predictions $\bm{Y}^{{\rm pred},1}_{\rm vot},\bm{Y}^{{\rm pred},2}_{\rm vot},\cdots,\bm{Y}^{{\rm pred},N}_{\rm vot}$ from each input point $\bm{x}_1,\bm{x}_2,\cdots,\bm{x}_N$ and then uses a point-to-point voting scheme to get the final estimation.
Specifically, we can regress an offset vector $\bm O_{ij}$ which predicts the vector beginning from input point $\bm x_i$ and ending at node $\bm y_j$. During inference, the $\bm y^{{\rm pred},i}_{j}$ can be calculated by adding the estimated $\bm O_{ij}^{\rm pred}$ to $\bm{x}_i$ and then we can apply a voting scheme among the predictions $\bm Y^{{\rm pred},i}_{\rm vot}$ of all input points to get the final $\bm{Y}^{\rm pred}_{\rm vot}$.

\begin{figure} [tb]
  \centering 
    \includegraphics[width=6cm]{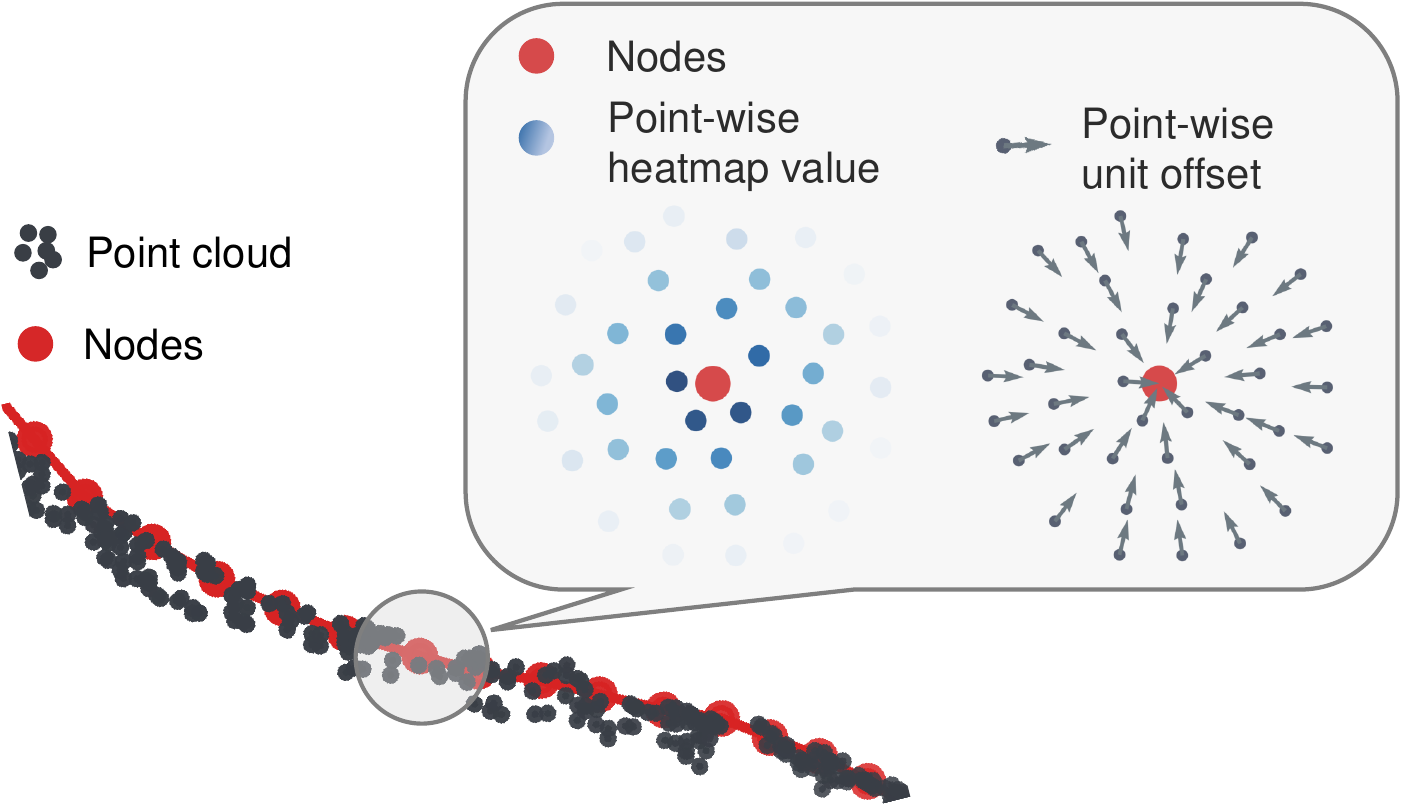} 
  \vspace{-2mm}
  \caption{Demonstration of predicted point-wise heatmap value and unit offset vector. Considering the neighborhood of one node, the points closer to it will have a higher heatmap value (visualized as deeper color), and the unit offset vector represents the normalized direction from the input point to the desired node.}
  \label{fig:voting}
    \vspace{-2mm}
\end{figure}

Similar to \cite{ge2018point}, we decompose the point-wise offset vector $\bm O_{ij}$ into a heatmap value $H_{ij}$ for the distance from $\bm x_i$ to $\bm y_j$ and a unit offset vector $\bm U_{ij}$ for the direction, as illustrated in Fig. \ref{fig:voting}, which makes network training easier. We also restrict the ground truth value of such a point-wise estimation inside the neighborhood of the desired node to exclude noisy estimations of points far away. Given a neighborhood radius $r$, the ground-truth heatmap value $H_{ij}^{\rm gt}$ is defined as 
\begin{equation}\label{Hij}
    H_{ij}^{\rm gt}=\left\{
\begin{array}{ccl}
1-\|\bm x_i-\bm y_j\|/r      &  , &    \|\bm x_i-\bm y_j\| <r,\\
0     & , & \|\bm x_i-\bm y_j\| \geq r,
\end{array} \right.
\end{equation}
and the ground-truth unit offset vector $\bm U_{ij}^{\rm gt}$ is defined as
\begin{equation}\label{Uij}
    \bm U_{ij}^{\rm gt}=\left\{
\begin{array}{ccl}
(\bm y_j-\bm x_i)/\|\bm x_i-\bm y_j\| & , & \|\bm x_i-\bm y_j\| <r,\\
0 & , & \|\bm x_i-\bm y_j\| \geq r,
\end{array} \right.
\end{equation}

We regress the point-wise heatmap $\bm H\in \mathbb{R}^{N \times M}$ and offset vector $\bm U\in \mathbb{R}^{N\times M \times 3}$ from the encoded latent feature $\bm F(\bm X)\in \mathbb{R}^{N\times C_{out}}$ using shared point-wise fully-connected layers $\bm{FC}_2$ and $\bm{FC}_3$, respectively.
The prediction of the heatmap and unit offset is denoted as
\begin{equation}
\bm H^{\rm pred} = {\rm Sigmoid} \left( \bm{FC}_2(\bm{F}(\bm X)) \right),
\end{equation}
\begin{equation}
\bm U^{\rm pred} = {\rm Normalized} \left( \bm{FC}_3(\bm{F}(\bm X)) \right),
\end{equation}
where the Sigmoid activation function and vector normalization are used for regulating the predicted heatmap value $\in [0, 1]$ and the unit offset, respectively.
For the point-to-point voting method, the training loss is
\begin{equation}
    L_{\rm vot}= \frac{1}{N} \sum_{i=1}^N \sum_{j=1}^M \left[ (H^{\rm pred}_{ij} - H^{\rm gt}_{ij})^2 + \|\bm U^{\rm pred}_{ij} - \bm U^{\rm gt}_{ij}\|^2\right].
\end{equation}
The overall training loss for the whole network can be formulated as: $L_{\rm tot} = L_{\rm reg} + \alpha \,  L_{\rm vot}$, where $\alpha$ is a pre-defined coefficient.

As for the inference, the point-wise estimation for node $\bm y_j$ from input point $\bm x_i$ is obtained as 
\begin{equation}
\bm y^{{\rm pred},i}_j = r\,(1-H^{\rm pred}_{ij})\,\bm U^{\rm pred}_{ij} + \bm x_i.
\end{equation}
We assume that the input point closer to the desired node predicts more accurate heatmap and unit offset, so we also use the heatmap value $H^{\rm pred}_{ij}$ as the confidence of the prediction $\bm y^{{\rm pred},i}_{j}$. 
To eliminate the influence of unconfident predictions, we choose input points with the highest $K$ headmap value for the $j^{\rm th}$ node to calculate the final estimation as
\begin{equation}
\bm y^{\rm pred}_j = \left( \sum_{i \in \mathcal{K}} H^{\rm pred}_{ij}\bm y^{{\rm pred},i}_j \right)/\sum_{i \in \mathcal{K}} H^{\rm pred}_{ij},
\end{equation}
where the indexes of the $K$ chosen points form the set $\mathcal{K}$.

Experiment results show that the point-to-point voting scheme can produce precise state estimations, suggesting that it mainly focuses on local regions and learns the local characteristic for precise estimation well. 
However, its architecture determines that if there are no enough input points in the local neighborhood of a node because of heavy occlusions, the prediction of the occluded part will be significantly inaccurate (also shown in Fig. \ref{fig:exp_sim_rvf}).

\subsection{Fusion of the Two Branches}

\begin{figure} [tb]
  \centering 
    \includegraphics[width=8.5cm]{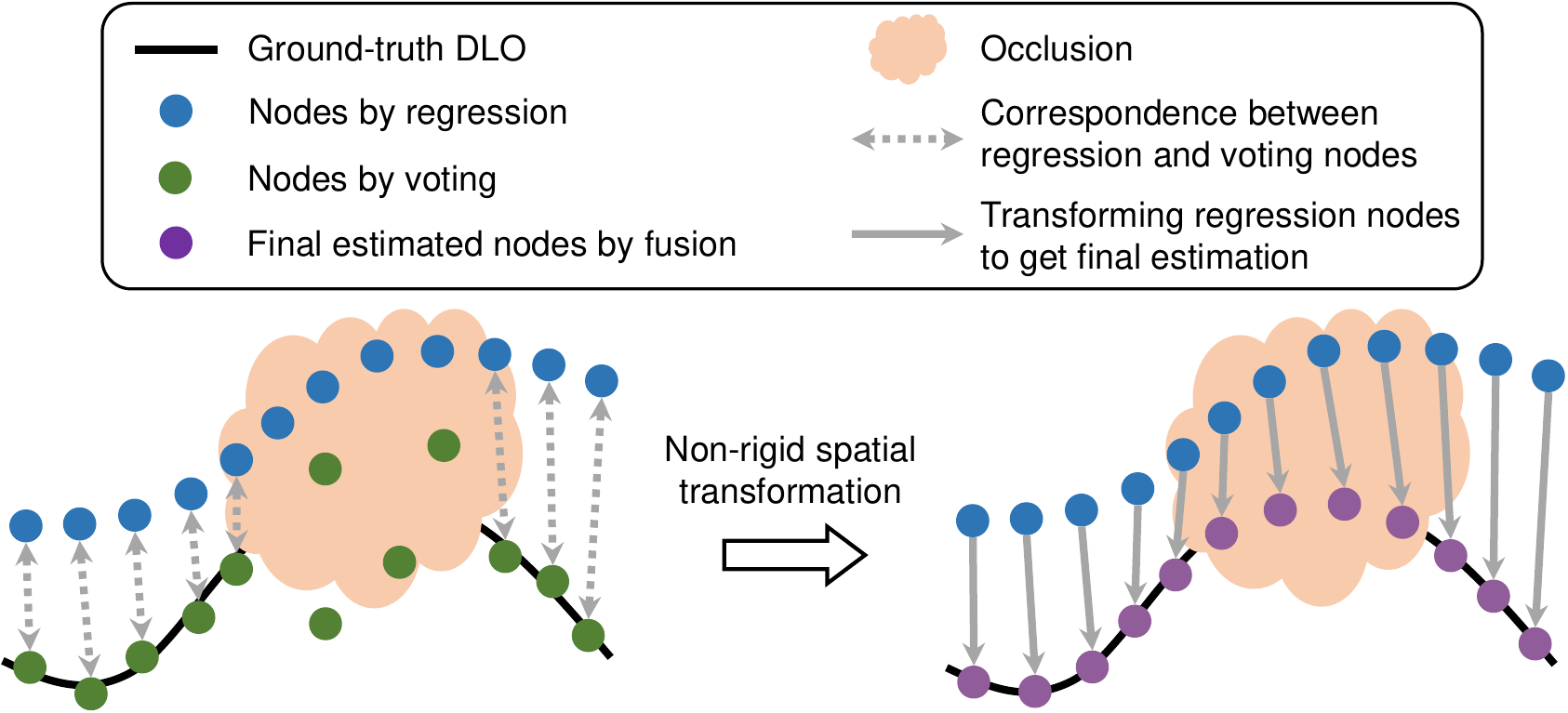} 
  \caption{Illustration of the non-rigid registration-based fusion module. We first identify and exclude the unreliable nodes estimated by voting (green points inside occlusion region). Then, a non-rigid spatial transformation is estimated by applying our modified CPD registration algorithm from the smooth but imprecise regression results (blue points in the left sub-picture) to the precise reliable voting results. Finally, the total sequence estimated by regression is transformed to get the final node sequence (purple points) using this non-rigid transformation.}
  \label{fig:fusion}
    \vspace{-2mm}
\end{figure}

To leverage the advantages of both two branches and achieve occlusion-robust state estimations, we further introduce a non-rigid registration-based fusion module.
With the known correspondence between the regression and voting results, we aim to estimate a non-rigid transformation from the smooth but imprecise regression results to the accurate unoccluded voting results, as shown in Fig. \ref{fig:fusion}.
Note that here we only select the subset of unoccluded nodes from the regression and voting results to exclude unreliable estimations for registration. 
Then, we transform all the nodes in regression results using the estimated transformation to obtain the final estimations whose unoccluded parts are almost as accurate as the voting results and occluded parts are filled up by the transformed smooth regression results. Details of each step are described as follows:  

\subsubsection{Select the unoccluded node subset}
Firstly, we have to recognize which node lies in the missing part of the input point cloud. 
We define the visible possibility  of the considered node $\bm y_j$ as the max value of the heatmap among all input points:


The nodes whose $p_j$ is greater than a pre-defined threshold $T \in [0, 1]$ are regarded as unoccluded parts. 
Denoting the visible possibility of all nodes $\bm Y$ as $\bm P$, the subsets of unoccluded nodes from the regression and voting results for the following non-rigid registration (simplified as $\textit{nrr}$) are
\begin{equation}\label{regpoints}
\begin{aligned}
    \bm Y_{\rm vot}^{\rm nrr} = \bm Y_{\rm vot}^{\rm pred} \,[ \,\bm P > T \,],\\
    \bm Y_{\rm reg}^{\rm nrr} = \bm Y_{\rm reg}^{\rm pred} \,[ \,\bm P > T \,].
\end{aligned}
\end{equation}

\subsubsection{Estimate the non-rigid transformation}
We utilize a modified \textit{Coherent Point Drift} (CPD) algorithm\cite{myronenko2010point}
to estimate the non-rigid transformation with known correspondences.
The classical CPD formulates registration as a GMM problem and regards source points ($\bm Y_{\rm reg}^{\rm nrr}$ for us) as the centroids of Gaussian model from which target points ($\bm Y_{\rm vot}^{\rm nrr}$) are sampled. For non-rigid registration, CPD ensures the coherent motion of these centroids by representing the non-linear spatial transformation as
\begin{equation}
    \mathcal{T}(\bm Y_{\rm reg}^{\rm nrr}) = \bm Y_{\rm reg}^{\rm nrr} + \bm G(\bm Y_{\rm reg}^{\rm nrr})\, \bm W,
\end{equation}
where $\bm G(\cdot)\bm W$ represents the displacement function as a Gaussian Radius Basis Function Network. The elements of kernel matrix $\bm G$ with a parameter $\beta$ are
\begin{equation}
    G_{ij}(\bm Z) = \exp(-\|\bm z_i-\bm z_j\|^2/2\beta^2),
\end{equation}
and $\bm W\in \mathbb{R}^{N_z\times D}$ ($N_z$ as the number of points in $\bm Z$, and $D=3$ for this 3-D task) is the weight matrix.

For general-defined non-rigid registration problem, estimating the correspondence between two point sets is very challenging, and CPD uses the EM algorithm to iteratively update the correspondence probability matrix, $\sigma^2$ of the Gaussian model, and the weight matrix $M$ until convergence. 
In the E-step, CPD fixes the current estimation of $\sigma^2$ and $\bm W$ to estimate more accurate correspondences while $\sigma^2$ and $\bm W$ are later optimized in the M-step. 
However, the correspondence of our source $Y_{\rm reg}^{\rm nrr}$ and target $Y_{\rm vot}^{\rm nrr}$ has been given naturally by the order of nodes in the sequence. Thus, there is no need to execute the E-step and we can directly update $\sigma^2$ and $\bm W$ iteratively.

Following Eq. (22) and Eq. (23) in \cite{myronenko2010point}, we can fix the correspondence probability matrix as an identity matrix and solve $\bm W$ and $\sigma^2$ using
\begin{equation}\label{eq:solve_w}
    (\bm G(\bm Y_{\rm reg}^{\rm nrr}) + \lambda \sigma^2 \bm I)\bm W = \bm Y_{\rm vot}^{\rm nrr} - \bm Y_{\rm reg}^{\rm nrr},
\end{equation}
\begin{equation}
\begin{aligned}
    \sigma^2 = & \frac{1}{D}( {\rm tr} ((\bm Y_{\rm vot}^{\rm nrr})^\transpose{} \bm Y_{\rm vot}^{\rm nrr}) - 2{\rm tr}((\bm Y_{\rm vot}^{\rm nrr})^\transpose{} \mathcal{T}(\bm Y_{\rm reg}^{\rm nrr})) \\
    &+ {\rm tr}(\mathcal{T}(\bm Y_{\rm reg}^{\rm nrr})^\transpose{}\mathcal{T}(\bm Y_{\rm reg}^{\rm nrr}))),
\end{aligned}
\end{equation}
where $\lambda$ is a parameter reflecting the level of smoothness regularization and $D=3$.


\subsubsection{Transform the whole set of regression results}
Finally, we apply the estimated non-rigid transformation to the whole regression node sequence, where
the displacement function is given by a Gaussian Radius Basis Function Network and the previously optimized weights $\bm W$.

 With all the regression nodes $\bm Y_{\rm reg}^{\rm pred}$ and our chosen subset $\bm Y_{\rm reg}^{\rm nrr}$, we reconstruct the kernel matrix $\bm G(\bm Y_{\rm reg}^{\rm pred}, \bm Y_{\rm reg}^{\rm nrr})$ whose elements are
\begin{equation}
    G_{ij}(\bm Z, \bm Y) = exp(-\|\bm z_i-\bm y_j\|^2/2\beta^2).
\end{equation}

The transformed regression node sequence are our final fused state estimations of two branches: 
\begin{equation}
    \bm Y_{\rm fus}^{\rm pred} = \mathcal{T}(\bm Y_{\rm reg}^{\rm pred}) = \bm Y_{\rm reg}^{\rm pred} + \bm G(\bm Y_{\rm reg}^{\rm pred}, \bm Y_{\rm reg}^{\rm nrr})\, \bm W.
\end{equation}

\section{Results}

\begin{table*}
\centering
\caption{Quantitative results of three methods with different levels of occlusion.}
\label{tab:exp_sim_diff_occlusion}
\setlength\tabcolsep{2pt}
\begin{tabular}{c|ccc|ccc|ccc|ccc} 
\toprule
Level of occlusion & \multicolumn{3}{c|}{No occlusion} & \multicolumn{3}{c|}{10\% occluded} & \multicolumn{3}{c|}{20\% occluded} & \multicolumn{3}{c}{40\% occluded} \\
Method & Regression & Voting & Fusion & Regression & Voting & Fusion & Regression & Voting & Fusion & Regression & Voting & Fusion \\ 
\hline
Errors of all nodes (mm) $\downarrow$ & 25.6 & \textbf{2.7} & \textbf{2.7} & 25.9 & 7.3 & \textbf{3.9} & 26.8 & 27.5 & \textbf{11.5} & 32.6 & 55.6 & \textbf{25.7} \\
Errors of unoccluded nodes~(mm) $\downarrow$ & 25.6 & \textbf{2.7} & \textbf{2.7} & 25.8 & \textbf{3.7} & 3.8 & 26.6 & \textbf{9.1} & 9.8 & 29.0 & \textbf{17.6} & 20.0 \\
Errors of occluded nodes (mm) $\downarrow$ & - & - & - & 26.6 & 39.6 & \textbf{5.2} & 27.9 & 101.1 & \textbf{18.2} & 35.2 & 112.3 & \textbf{33.4} \\
Uniformity (mm) $\downarrow$ & 1.1 & 1.1 & \textbf{0.9} & 1.10 & 17.5 & \textbf{1.0} & \textbf{1.1} & 21.3 & 1.7 & \textbf{1.1} & 22.6 & 2.6 \\
\bottomrule
\end{tabular}
\vspace{-3mm}
\end{table*}

\subsection{Data Collection and Model Training}

All the training data are generated in simulations for the convenience of getting the ground-truth node positions. We use the Unity3D \cite{unity} as the simulator and the Obi Rope package \cite{obi} for simulating DLOs, which is a unified particle physics that models DLOs as chains of oriented particles with stretch, shear, bend, and twist constraints.

During the data collection, RGB-D images captured by a camera and 3-D positions of the DLO particles in the camera frame are recorded per simulation second to generate corresponding point cloud data and ground-truth node positions. The two ends of the DLO are randomly moved using the same strategy as in our previous work \cite{yu2022global} which is efficient in covering different shapes. Visualizations of the data examples can be found in the supplemental video on our project website\footnote{\url{https://kangchen-lv.github.io/DLO-perception}}.
The DLO properties (lengths, radius, and stiffness) and camera poses are also randomized to improve the generalization ability of the model. 

We totally generate $25000$ frames of data (500 sequences, 50 frames in each sequence) and randomly separated them with a ratio of 0.8 and 0.2 for training and validation, respectively.
During the training, the input point cloud is augmented by random jittering with Gaussian noises, random rotation across each axis, and random occlusions. For batch training, we sample $N=1024$ points from the initial point cloud with the farthest point sampling (FPS) method and predict $M=50$ nodes from it.

In our implementation, we build up the PointNet++ encoder as the architecture for segmentation tasks in \cite{qi2017pointnet++}, which has 4 point set abstraction levels and 4 feature propagation levels.
As for hyper-parameters, we use the Adam optimizer with a initial learning rate 0.01, a weight decay 5e-4 and a batch size 32 to train the network for 200 epochs. The learning rate decays with a ratio 0.8 every 20 epochs. The training loss weight of two branches is $\alpha=1.0$. For point-to-point voting branch, we set $r=0.02$ and $K=64$. For the registration-based fusion module, we set $T=0.5$, $\lambda=0.25$ and $\beta=0.5$.



\begin{figure} [tb]
  \centering 
  \subfigure[]{ 
    \includegraphics[width=0.225\textwidth]{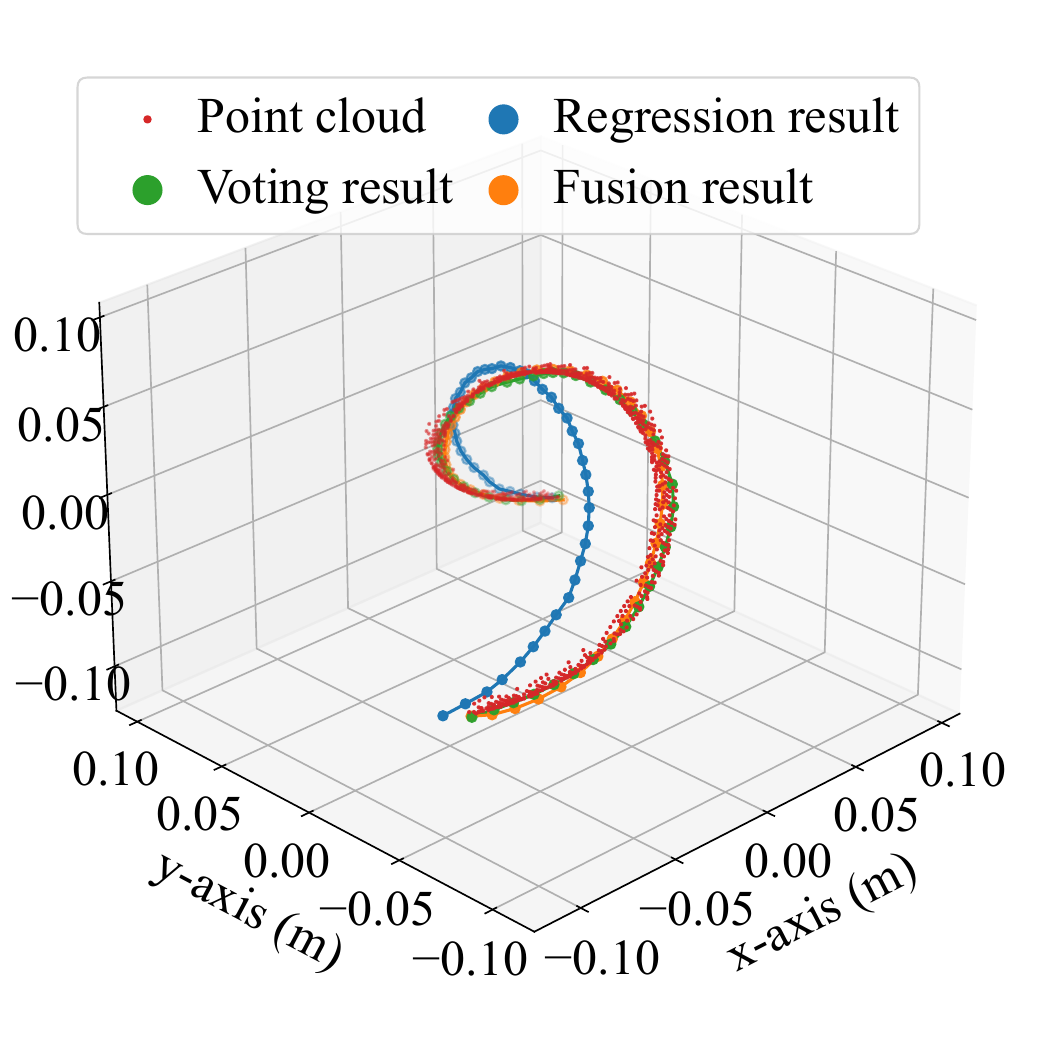} 
  } 
  \hspace{-0.5cm}
  \subfigure[]{ 
    \includegraphics[width=0.225\textwidth]{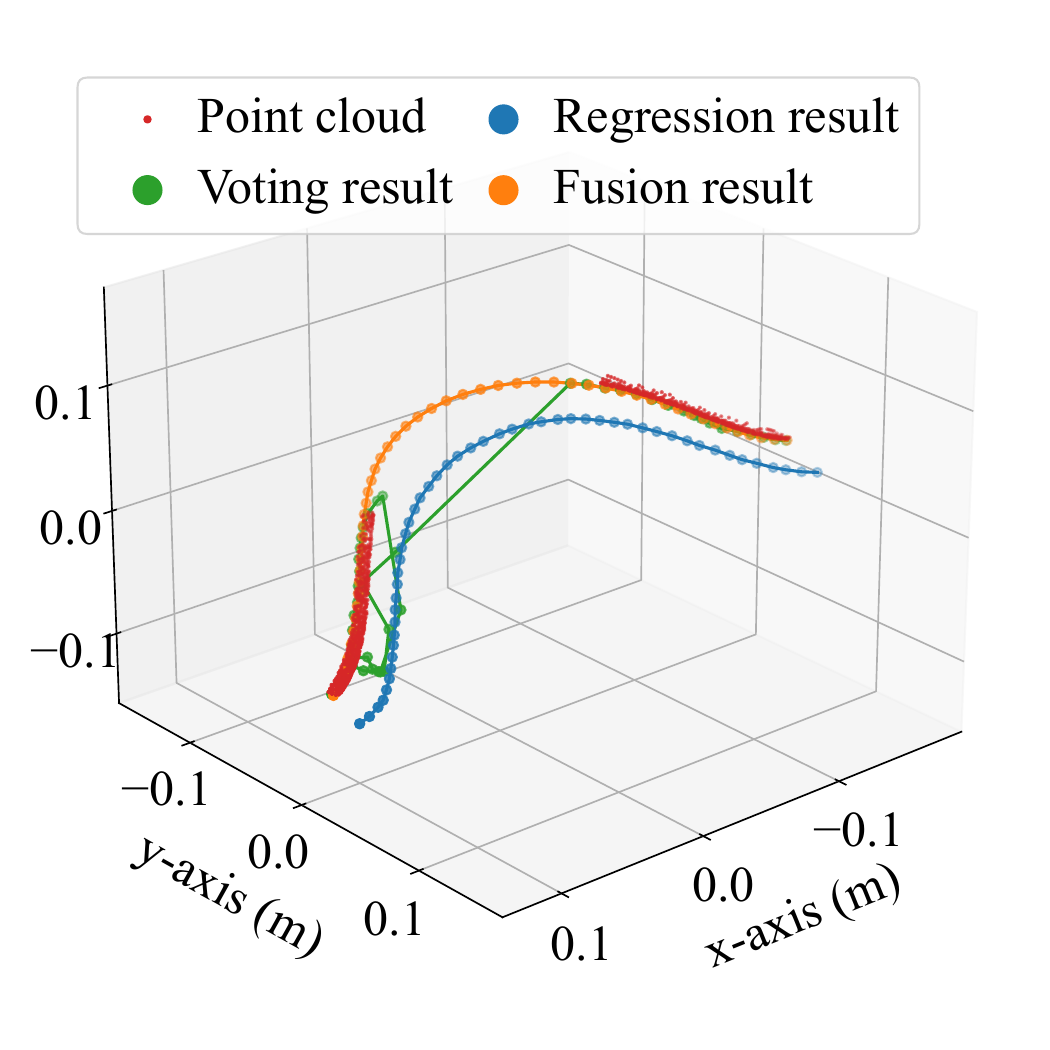} 
  }
  \vspace{-2mm}
  \caption{Visualization of the regression, voting and fusion results of  (a) unoccluded point cloud and (b) occluded point cloud.}
  \label{fig:exp_sim_rvf}
  \vspace{-2mm}
\end{figure}


\subsection{Simulation Results}

We adopt two metrics to evaluate the performance: a) errors of nodes, i.e., the average Euclidean distance between the estimated and ground-truth node positions; and b) uniformity, i.e., the standard deviation of the distances between every adjacent node which illustrates whether the estimated nodes are uniformly distributed.  

\begin{figure} [tb]
  \centering 
  \subfigure[20\% occluded]{ 
    \includegraphics[width=0.145\textwidth]{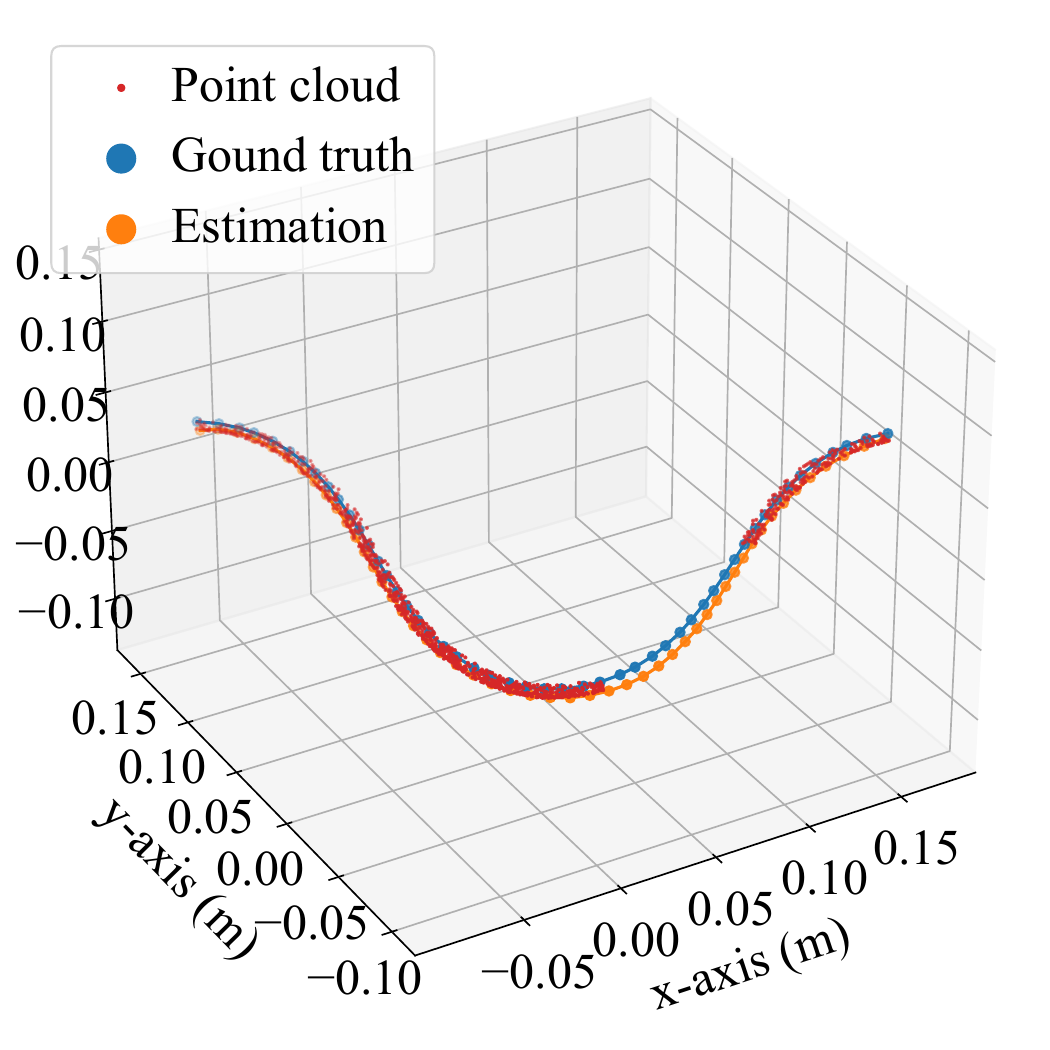}
  } 
  \hspace{-0.4cm}
  \subfigure[40\% occluded]{ 
    \includegraphics[width=0.145\textwidth]{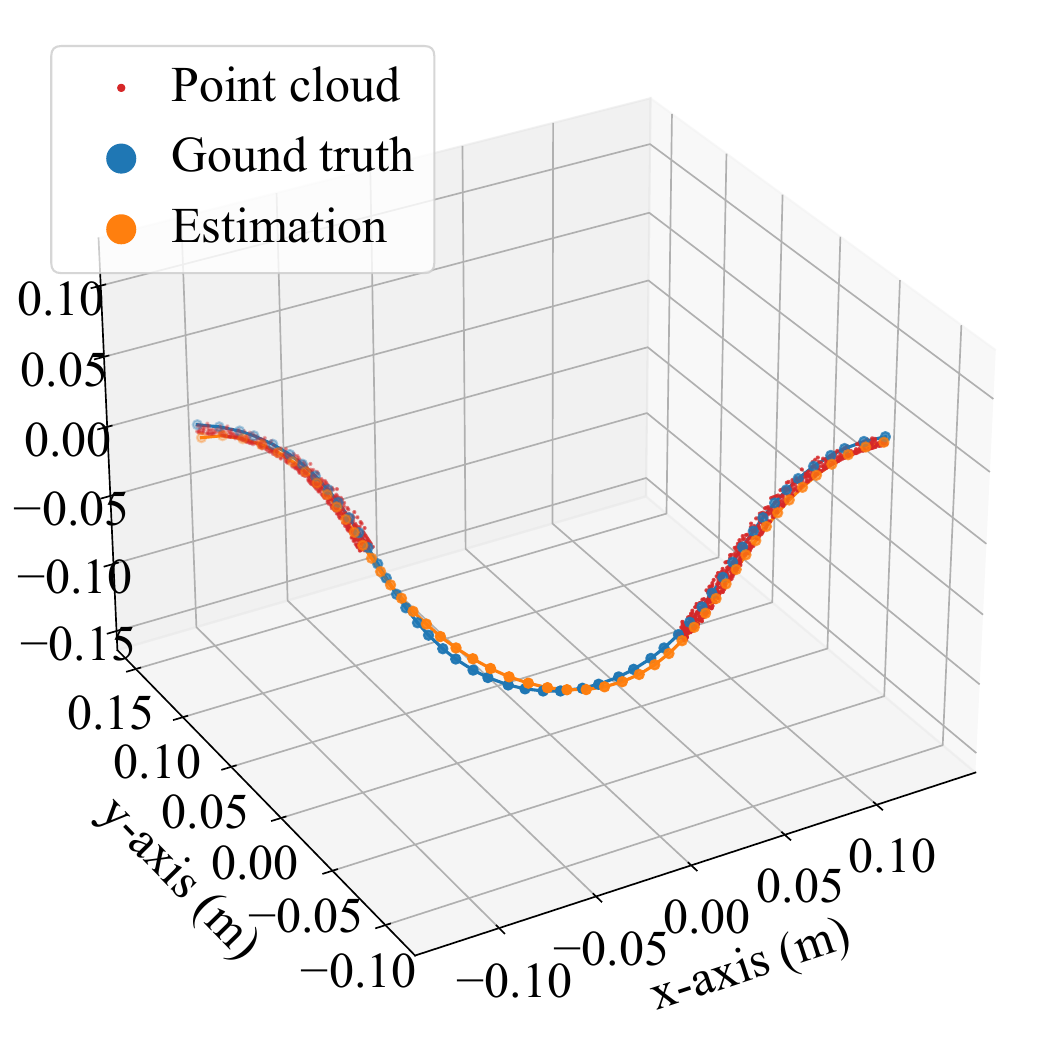}
  } 
  \hspace{-0.4cm}
  \subfigure[60\% occluded]{ 
    \label{fig:exp_sim_occlusion_60} 
    \includegraphics[width=0.145\textwidth]{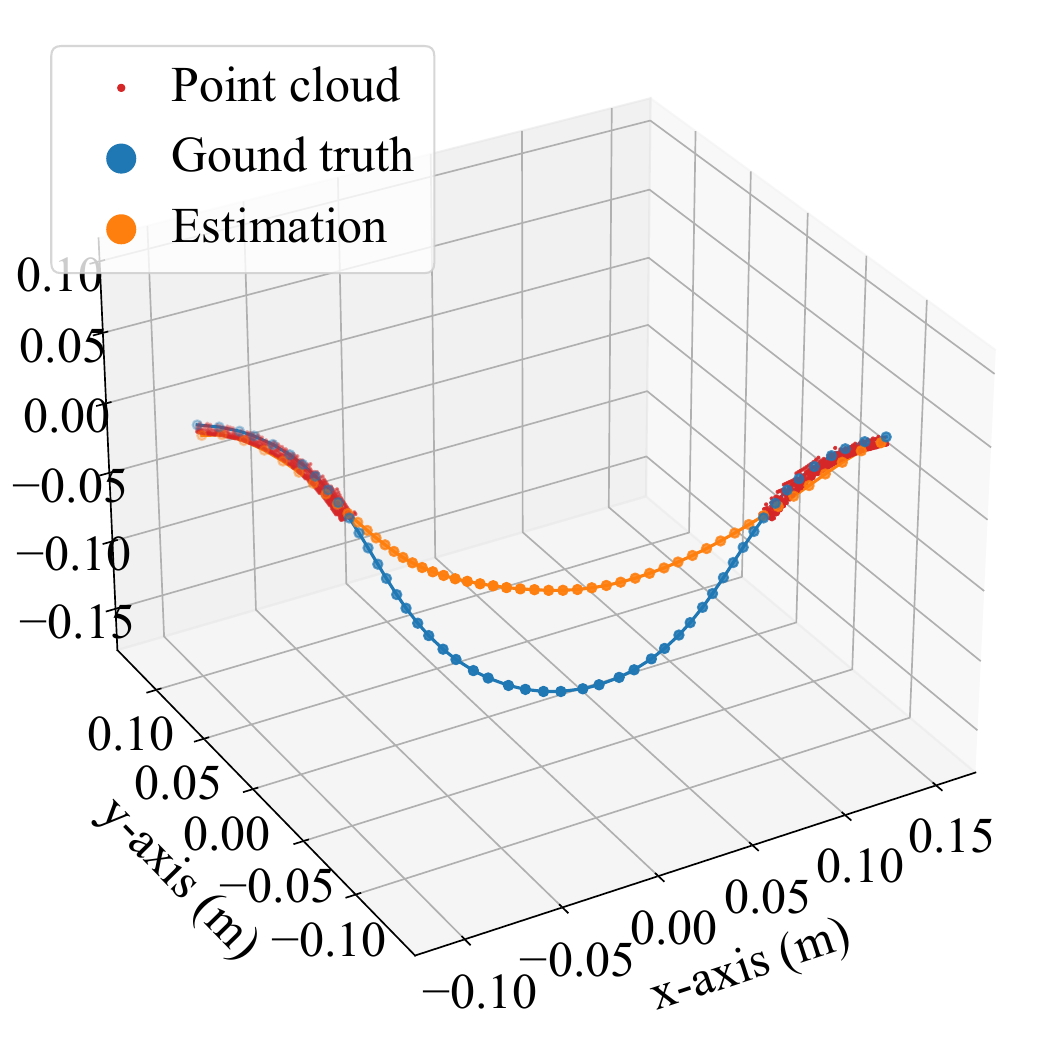}
  } 
  \vspace{-2mm}
  \caption{Estimation results of the same DLO under different occlusion ratio.}
  \label{fig:exp_sim_occlusion}
  \vspace{-3mm}
\end{figure}

\begin{figure} [tb]
  \centering 
  \subfigure[]{ 
    \label{fig:exp_sim_T} 
    \includegraphics[width=0.225\textwidth]{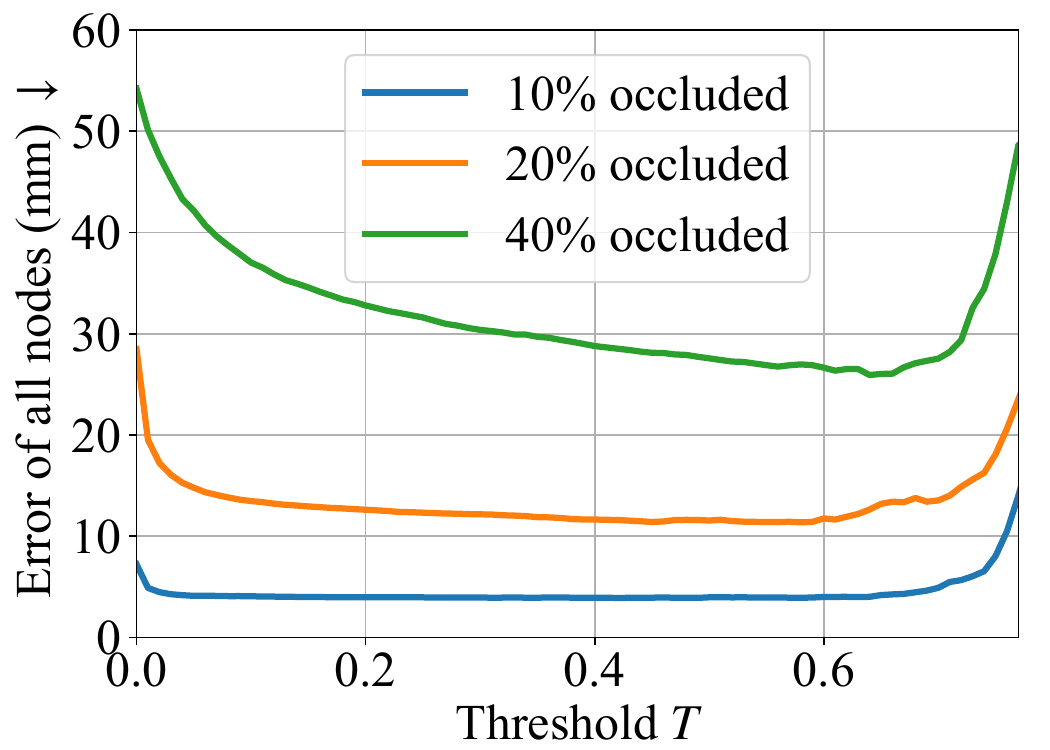} 
  } 
  \hspace{-0.2cm}
  \subfigure[]{ 
    \label{fig:exp_sim_noise} 
    \includegraphics[width=0.225\textwidth]{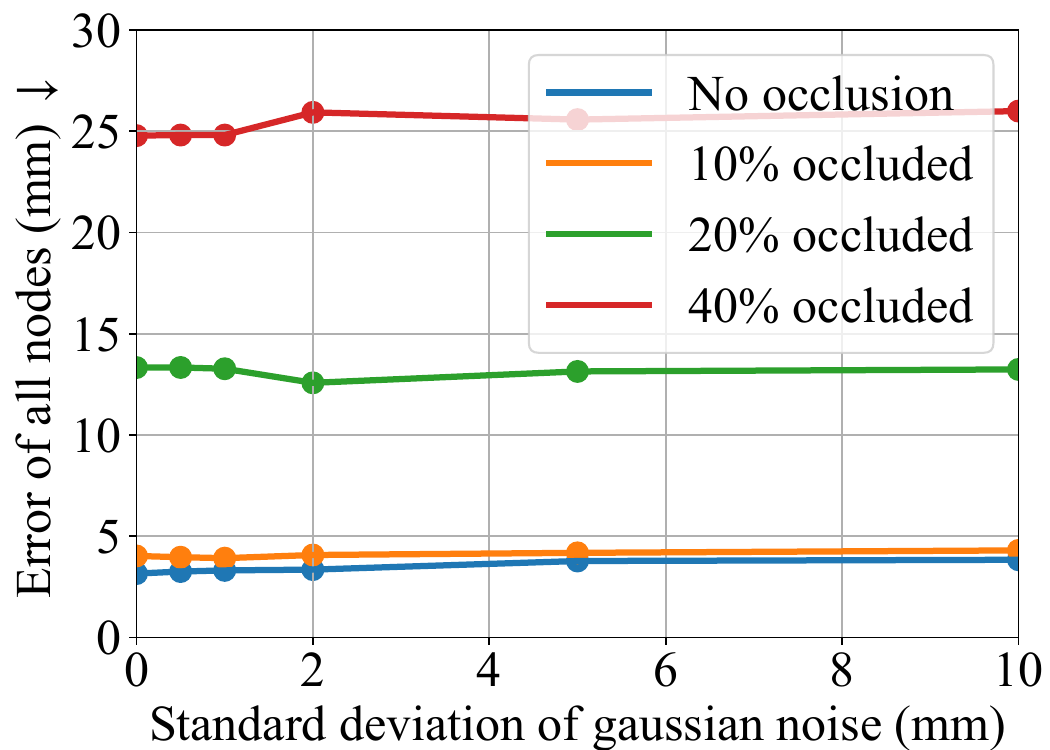} 
  }
  \vspace{-2mm}
  \caption{(a) The relationship between node error and the visible possibility threshold $T$. (b) The robustness of our method under point cloud noises.}
  \label{fig:exp_T_noise}
\end{figure}

\begin{figure*} [tb]
  \centering 
    \includegraphics[width=17.3cm]{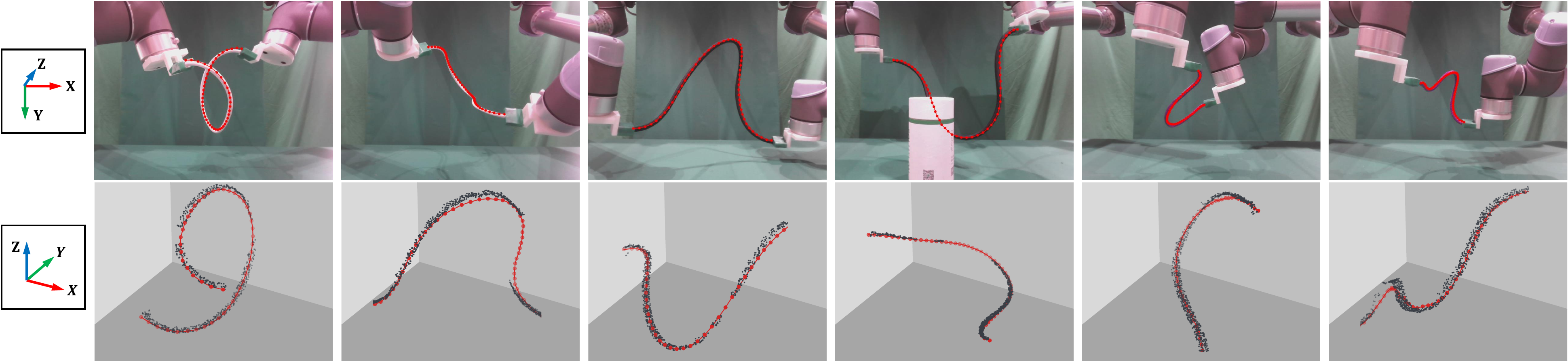} 
  \caption{State estimation results of three different real-world DLOs from occluded and fragmentary point clouds. The top row shows the raw RGB images and the reprojection of the estimated node sequence (in red color); and the bottom row shows the raw point clouds of the DLOs (in black color) and the 3-D positions of the estimated node sequence (in red color). Each column refers to a shape, in which the top image is in the common camera frame and the bottom point cloud is in a top-view frame to better illustrate the Z-axis shape.}
  \label{fig:exp_real_images}
    \vspace{-2mm}
\end{figure*}

\subsubsection{Self-comparisons}
We quantitatively evaluate the performance of \textit{End-to-End Regression}, \textit{Point-to-Point Voting} and \textit{Fusion} with different levels of occlusion, in which we further decompose the distance error of all nodes into those of unoccluded and occluded nodes to suggest the specific properties of these methods.
As shown in Table \ref{tab:exp_sim_diff_occlusion}, the errors of regression results are relatively large compared to other methods in unoccluded cases but increase very slowly as the level of occlusion rises. 
Besides, the regression results keep uniform under all settings, suggesting that it always estimates smooth global DLO shapes even with heavy occlusions.
In contrast, although the voting results are precise for complete point cloud input (with an error of $2.7$mm), 
the prediction accuracy and uniformity cannot be guaranteed for occluded cases (with an error of $112.3$mm for 40\% occluded input). Under all settings, our proposed fusion method is the most accurate and robust one, which owes to its ability to combine the advantages of both two branches to predict accurate positions close to the unoccluded voting results and maintain a smooth shape.
The uniformity is also well kept even in heavily-occluded scenarios.

Two visualized examples are shown in Fig. \ref{fig:exp_sim_rvf}. Note that the estimated nodes are connected one by one to illustrate their order in all visualizations. It can be seen that the regression results are robust against occlusion but imprecise, while the voting results are accurate outside occlusion but extremely unreliable for occluded parts.
However, our fusion method can accurately and robustly estimate the DLO state in both unoccluded and occluded scenarios.

\begin{table}
\centering
\caption{Ablation studies of different fusion methods.}
\label{tab:exp_sim_ablation}
\setlength\tabcolsep{3pt}
\begin{tabular}{c|c|cc} 
\toprule
\begin{tabular}[c]{@{}c@{}}Level of\\occlusion\end{tabular} & Fusion method & \begin{tabular}[c]{@{}c@{}}Errors of all\\nodes (mm)~$\downarrow$\end{tabular} & \begin{tabular}[c]{@{}c@{}}Uniformity \\(mm)~$\downarrow$\end{tabular} \\ 
\hline
\multirow{4}{*}{No occlusion} & Point Concate. & 15.9 & 1.2 \\
 & Latent Concate. & 18.0 & 1.2 \\
 & CPD & 16.5 & 1.2 \\
 & Ours & \textbf{2.7} & \textbf{0.9} \\ 
\hline
\multirow{4}{*}{20\% occluded} & Point Concate. & 18.3 & \textbf{1.2} \\
 & Latent Concate. & 20.7 & \textbf{1.2} \\
 & CPD & 19.4 & 1.7 \\
 & Ours & \textbf{11.5} & 1.7 \\
\bottomrule
\end{tabular}
\vspace{-3mm}
\end{table}

\subsubsection{Robustness and sensitivity analysis}
First, we test the relationship between the performance of our method and the occlusion level.
As shown in Fig.~\ref{fig:exp_sim_occlusion}, the occlusion ratio of point cloud is 20\%, 40\%, and 60\%, respectively. When the most of the DLO is occluded (such as Fig. \ref{fig:exp_sim_occlusion_60}), the accurate DLO state is very difficult to estimate since there are multiple possible shapes for the occluded part, but our predicted node sequence still forms a reasonable DLO shape given only the incomplete information of two DLO ends.

In addition, we investigate the sensitivity of our fusion module to the threshold $T$, which is for determining the visible nodes in Eq.~\ref{regpoints}, as shown in Fig.~\ref{fig:exp_sim_T}.
As $T$ increases, more unreliable voting nodes for occluded parts will be excluded from registration so that the distance error will decrease. However, too many valid nodes will also be excluded and the performance will be hurt severely with a high $T$ value.
We also conduct a test of the robustness against Gaussian noises on the input point cloud. As demonstrated in Fig. \ref{fig:exp_sim_noise}, despite that the standard deviation of noise is increasing, our performance still keeps stable.


\subsubsection{Ablation on fusion approaches}
Several fusion methods are compared with the proposed modified CPD-based fusion module. 
We design the baseline fusion methods as follows: a) Point Concate.: concatenating the voting and regression results and learning a refinement mapping with an MLP; b) Latent Concate.: concatenating the voting results and the global feature in regression branch to learn the refinement; c) CPD: the classical CPD algorithm\cite{myronenko2010point} with unknown correspondences. 
Experimental results in Table \ref{tab:exp_sim_ablation} suggest that our fusion method outperforms all baselines in both occluded or unoccluded scenarios. 
Note that the first two baselines perform similar to the regression module so that their uniformity under occlusion is good.

\subsection{Real-World Experiments}


\begin{figure} [tb]
  \centering 
    \includegraphics[width=8.4cm]{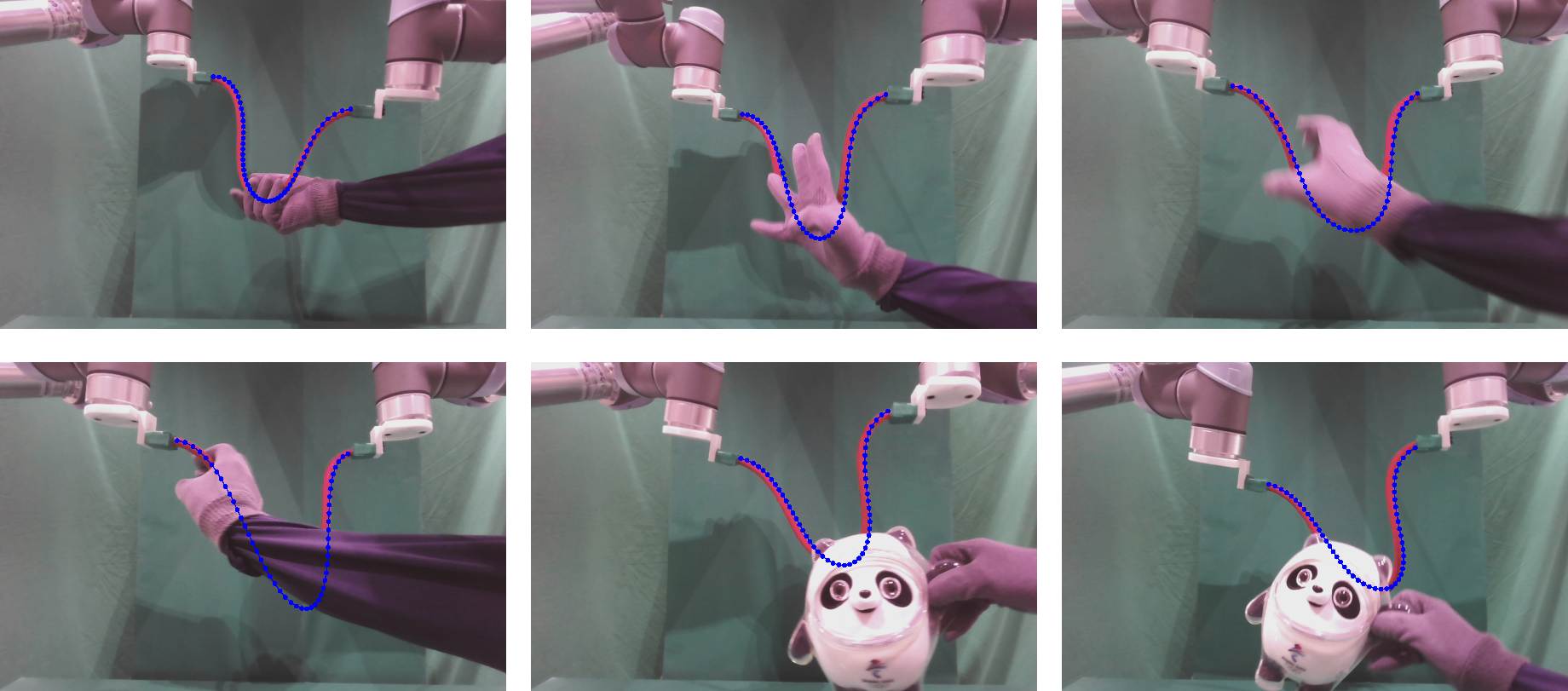} 
  \vspace{-1mm}
  \caption{State estimation results of a moving DLO under partial occlusions by hands or the "Bing Dwen Dwen". The blue points refer to the reprojection results of the estimated 3-D node positions.}
  \label{fig:exp_real_sequence}
    \vspace{-3mm}
\end{figure}


We choose three DLOs of different lengths (0.5, 0.7, and 0.3m, respectively) and different materials to examine the generalization ability of our method in real-world applications. The two ends of DLOs are rigidly grasped by dual robot arms and deformed to various complex shapes. 
The DLO point clouds are first segmented from the background via color thresholding in RGB space and then fed into our two-branch model.
Results in Fig. \ref{fig:exp_real_images} illustrates that our method can be directly applied to estimate the real-world DLO state with small sim-to-real gaps. 
Even in some cases with self-intersection or occlusion by obstacles, our state estimations are still smooth and precise enough.
Experiments on estimating the state of a moving DLO from each frame in a dynamic sequence (see Fig. \ref{fig:exp_real_sequence}) also suggest the robustness of our method amidst heavy occlusions.

\begin{figure} [tb]
  \centering 
    \includegraphics[width=8.4cm]{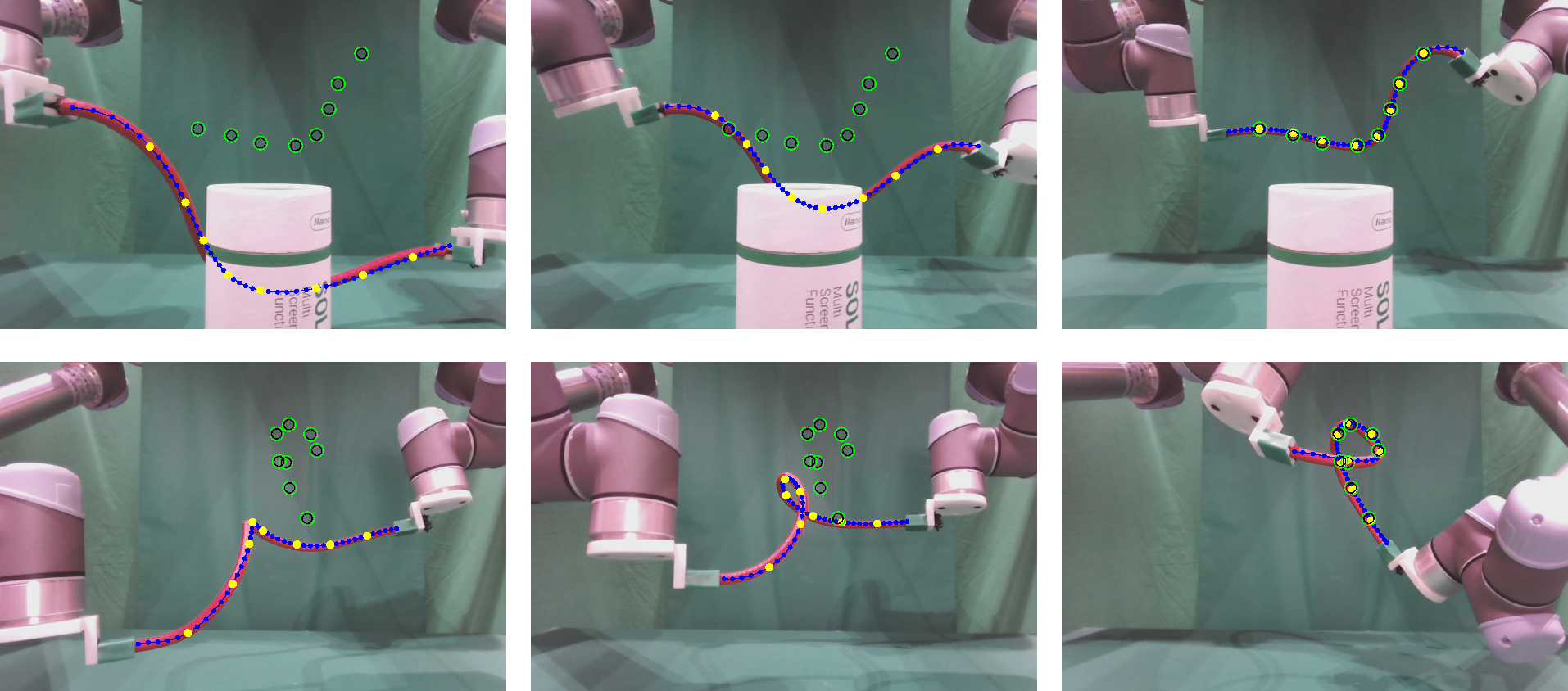} 
  \vspace{-1mm}
  \caption{Applications in downstream DLO shape control task. The blue, yellow, and green points represent the reprojected 3-D node positions, selected control nodes and target positions, respectively.}
  \label{fig:exp_manipulation}
    \vspace{-3mm}
\end{figure}

We also integrate our method into the DLO shape control task as the front-end perception module. As shown in Fig. \ref{fig:exp_manipulation}, a uniformly-distributed subset of estimated nodes (8 yellow points) is chosen to be controlled to achieve the target positions (corresponding green points) and form the desired DLO shape. The controller is based on our previous work \cite{yu2022global} and our method can achieve real-time performance on a GeForce RTX 2060 GPU (average computation time per frame is 34ms). In the presence of occlusions or self-intersections at both initial and middle stage of the manipulation process, our method can steadily output precise DLO states and finally achieve the target positions, which cannot be realized by the existing pure-tracking methods.

\section{Conclusions}
In this work, we propose a learning-based method to robustly estimate the 3-D states of DLOs from single-frame point clouds even with heavy occlusions. We use a sequence of ordered nodes as the state representation of DLOs and design a two-branch architecture to estimate it. The two branches are encouraged to utilize the global or local geometry information respectively and their estimations are combined by a fusion module later to get the final output. The simulation and real-world experimental results demonstrate our method can guarantee an accurate and smooth shape of DLOs in both occluded and unoccluded cases with high generalization capability for real-world applications. In the future, we will be devoted to incorporating the temporal information into the framework to get smoother tracking results across a long-term point cloud sequence. 

\clearpage
\bibliographystyle{IEEEtran}
\bibliography{ref}

\end{document}